\documentclass[runningheads]{llncs}
\usepackage{amsmath}
\usepackage{amssymb}
\usepackage{bbm}
\usepackage{algorithm}
\usepackage{algpseudocode}
\usepackage{graphicx}
\usepackage{subcaption}
\usepackage{multirow}
\usepackage{rotating}
\usepackage{comment}
\graphicspath{{./imgs/}}

\begin{document}
\title{On the Importance of Regularisation \& Auxiliary Information in OOD Detection\thanks{This research is supported by Sience Foundation Ireland (SFI) under Grant number SFI/12/RC/2289\_P2.}}
\titlerunning{Regularisation \& Auxiliary Infomation in OOD Detection}
%
\author{
John Mitros\orcidID{0000-0002-0189-2130} \and
Brian Mac Namee\orcidID{0000-0001-7842-1516}
}
\authorrunning{J. Mitros et al.}
%
\institute{
University College Dublin, IR \\
School of Computer Science \\
\email{ioanni.mitro@ucdconnect.ie, brian.macnamee@ucd.ie}
}
\maketitle              
\begin{abstract}
Neural networks are often utilised in critical domain applications (e.g. self-driving cars, financial markets, and aerospace engineering), even though they exhibit overconfident predictions for ambiguous inputs. This deficiency demonstrates a fundamental flaw indicating that neural networks often overfit on spurious correlations. To address this problem in this work we present two novel objectives that improve the ability of a network to detect out-of-distribution samples and therefore avoid overconfident predictions for ambiguous inputs.  We empirically demonstrate that our methods outperform the baseline and perform better than the majority of existing approaches while still maintaining a competitive performance against the rest. Additionally, we empirically demonstrate the robustness of our approach against common corruptions and demonstrate the importance of regularisation and auxiliary information in out-of-distribution detection.

\end{abstract}
\section{Introduction}
Out of distribution (OOD) detection is becoming more important as machine learning solutions are developed for critical applications (e.g.~self-driving cars, financial markets, and aerospace engineering), and especially in evaluating the robustness of deployed models. The main goal of OOD is to equip a classifier with the ability to provide stable, consistent and low confidence predictions for data points that might be far away from the in-distribution (ID) training data. This is often referred to as the capacity of the model to generalise. 

A central assumption in statistical learning theory~\cite{vapnik2013nature,vapnik1999overview,bousquet2003introduction,vonLuxburg2011statistical,mendelson2003few} states that the train and test set $(x_{i=1}^{n}, y_{i=1}^{n})$, are generated independently and identically distributed (IID) from a distribution $P$, such that data points are usually assigned randomly to either train or test set.Unfortunately, this assumption fails to assess whether the model has learned to properly generalise to new unseen data or has simply overfit to irrelevant factors (e.g., backgrounds in image recognition task) that might be spuriously correlated with the correct label due to shortcut learning~\cite{geirhos2020shortcut}. Numerous  methods have been proposed to mitigate this deficiency and improve OOD detection~\cite{lee2017a,liang2018enhancing,shalev2018,lee2018,ren2019likelihood,nandy2020,lee2020_bayesian_maml_ood} within supervised, semi-supervised, and unsupervised learning~\cite{bulusu2020}, including discriminative~\cite{schulam2019,ovadia2019,hsu2020} and generative~\cite{choi2018,liu2020,li2020,zisselman2020} models. In addition, some methods cast the OOD problem as binary classification~\cite{shafaei2018} with alternative approaches relying either only on ID data~\cite{yonatan2019} or both ID and OOD data~\cite{hendrycks2018_outlier_exposure} during training.

Inspired by recent progress in contrastive learning~\cite{oord2018,goyal2019,henaff2020} we propose two novel objectives for OOD detection and demonstrate empirically that our method not only is competitive with existing approaches but it also outperforms some of them in most occasions. Additionally, we empirically study the role of regularisation in OOD detection and robust classification. \\

In this work we investigate the following questions:
\begin{itemize}
    \item Can  contrastive learning improve OOD detection in neural networks?
    \item What is the role of explicit regularisation in OOD detection, and does additional regularisation improve or degrade OOD detection?
\end{itemize}
The main contributions of this work are:
\begin{itemize}
  \item A novel objective function based on the cosine angle between the ID and OOD data.
  \item A novel objective function inspired by prior work on margin and ranking objectives utilising the cosine angle between ID and OOD data  as well as additional explicit regularisation.
\end{itemize}

\section{Related Work}

In this section we describe existing work on the OOD problem, and the objectives used in recent approaches based on contrastive learning. 

\subsection{Out of Distribution Detection}

Early attempts at OOD detection~\cite{hendrycks2017} used the maximum softmax probability as an indicator to identify OOD samples, while alternative approaches such as ODIN~\cite{liang2018enhancing} use adversarially perturbed samples while computing the softmax with high temperature during training. Furthermore, the Mahalanobis detector~\cite{lee2018} fits a Gaussian distribution to the activation of the last layer of a neural network and performs OOD by measuring the Mahalanobis distance from the inputs to the ID data.

In addition, methods explicitly trained to output uniform distribution over ID perturbed samples, usually resemble techniques simulating OOD inputs from a GAN~\cite{lee2017a}, or utilising auxiliary information (e.g. additional datasets) as outliers~\cite{hendrycks2018_outlier_exposure}. Finally, there exist approaches relying on averaging predictions of randomly initialised, independently trained, neural networks, either in a discriminative~\cite{lakshminarayanan2017} or generative~\cite{choi2018} approach.

A naturally occurring question is: ``\textit{\textbf{What do these methods have in commmon?}}'' To the best of our knowledge the majority of techniques proposed to tackle the OOD problem can be attributed to one of the following categories: \textit{optimisation}, \textit{regularisation}, or \textit{sampling}---or a combination of the three.

\subsection{Objective Functions in Machine Learning}

Most objectives adopted today in machine learning (e.g. cross-entropy, mean squared error, and  log-likelihood) have a single goal, to induce a cost in order for the underlying model to directly learn a label, a value, or a set of values from a specific input. In contrast, ranking objectives strive to predict similarities (i.e. relative distances) between inputs, thus the underlying task is often identified and referred to as metric learning. The key idea is to employ a metric function (e.g. Euclidean distance) in order to obtain a similarity score between inputs embedded in a latent feature space, where the score should be small for similar inputs and large otherwise. One such example is SimCLR~\cite{chen2020} that maximises the agreement in latent representations via a contrastive objective between pairs of inputs.

Let $\operatorname{cos}(\mathbf{u}, \mathbf{v})=\mathbf{u}^{\top} \mathbf{v} /\Vert\mathbf{u}\Vert \Vert\mathbf{v}\Vert$ define a similarity score indicated by the cosine angle between vectors $\mathbf{u}$ and $\mathbf{v}$. This is utilised in the SimCLR objective. Given a pair of distinct latent features $(\mathbf{z}_{i}, \mathbf{z}_{j})$, such that $\mathbf{z}_{i,j} = f_{\theta}(t_{i,j}(\mathbf{x})), \forall \mathbf{x}\in\mathcal{X}$, with augmentation operations $t_{i}, t_{j}\sim\mathcal{T}$, such that $t_{i}\neq t_{j}$, and temperature scaling $\tau$ then the objective is formulated as:

\begin{align}
\label{eqn:simclr_objective}
 L(\mathbf{z}_{i}, \mathbf{z}_{j})&=-\log \frac{\exp \left(\operatorname{cos}\left(\mathbf{z}_{i}, \mathbf{z}_{j}\right) / \tau\right)}{\sum_{j=1}^{2n} \mathbbm{1}_{[j \neq i]} \exp \left(\operatorname{cos}\left(\mathbf{z}_{i}, \mathbf{z}_{j}\right) / \tau\right)}
\end{align}

Although ranking objectives (e.g. pairwise, triplet, etc.) might differ with regards to the number of inputs they operate on (e.g. pairs or triplets), nevertheless, their main concept is to learn a similarity/dissimilarity metric, (e.g. $\ell^{p}$-norm) on latent representations $(\mathbf{z}_{i}, \mathbf{z}_{j})$ such that $d = \Vert \mathbf{z}_{i}, \mathbf{z}_{j}\Vert_{2}$ for $j=\left\{1,\ldots,n\right\}$. Given a set of inputs $\left\{\mathbf{x},\mathbf{x}_{1}^{+},\mathbf{x}_{2}^{-}\ldots,\mathbf{x}_{n}^{-}\right\}$ with one positive and $n-1$ negative samples, where $\mathbf{x}$ represents an anchor sample, $\mathbf{x}_{i=1}^{+}$ a positve sample and $\mathbf{x}_{j=2,\ldots,n}^{-}$ a negative sample, with $y=\left\{0, 1\right\}$ being the labels. Then, a \textbf{pairwise ranking objective} strives to learn representations with small distance $d$ between positive pairs $(\mathbf{x}, \mathbf{x}_{i}^{+})$ and greater than a margin $\gamma$ for negative pairs $(\mathbf{x}, \mathbf{x}_{j}^{-})$ such that.
\begin{align}
\label{eqn:pairwise_ranking_objective}
L(\mathbf{x}_{i}, \mathbf{x}_{j}, y) &=
    \begin{cases}
        y\,d(\mathbf{z}, \mathbf{z}^{+}) & \text { if } (\mathbf{x}, \mathbf{x}_{i}) \text { is a positive pair } \\
        (1 - y)\,\max(0, \gamma - d(\mathbf{z}, \mathbf{z}^{-})) & \text { if } (\mathbf{x}, \mathbf{x}_{j}) \text { is a negative pair}
    \end{cases}
\end{align}

Instead of pairs the \textbf{triplet ranking objective} uses triplets $\left\{\ldots,\mathbf{x}, \mathbf{x}^{-}, \mathbf{x}^{+},\ldots\right\}$. We have an anchor $\mathbf{x}$, a positive $\mathbf{x}^{+}$, and a negative $\mathbf{x}^{-}$ sample, instead of pairs of positive $(\mathbf{x}, \mathbf{x}^{+})$ and negative $(\mathbf{x}, \mathbf{x}^{-})$ samples as illustrated in Equation~\ref{eqn:pairwise_ranking_objective}. The goal is to learn representations with greater distance between the anchor and the negative sample $d(\mathbf{z}, \mathbf{z}^{-})$ than between the anchor and the positive sample $d(\mathbf{z}, \mathbf{z}^{+})$. The final triplet objective is formulated as:
\begin{align}
L(\mathbf{x}, \mathbf{x}^{+}, \mathbf{x}^{-})
&= \max(0, \gamma + d(\mathbf{z}, \mathbf{z}^{+}) - d(\mathbf{z}, \mathbf{z}^{-})) \\
&=\begin{cases} \nonumber
    \text{Easy triplets:} & \text{ if } d(\mathbf{z}, \mathbf{z}^{-}) > d(\mathbf{z}, \mathbf{z}^{+}) + \gamma\\
    \text{Semi-hard triplets:} & \text{ if } d(\mathbf{z}, \mathbf{z}^{+}) < d(\mathbf{z}, \mathbf{z}^{-}) <  d(\mathbf{z}, \mathbf{z}^{+}) + \gamma \\
    \text{Hard triplets:} & \text{ if } d(\mathbf{z}, \mathbf{z}^{-}) < d(\mathbf{z}, \mathbf{z}^{+})
\end{cases}
\end{align}

\section{Objective Functions for OOD Detection}

The first contribution that this article makes to OOD detection is a novel objective based on the cosine similarity between ID and OOD predictions. Let $(\mathbf{x}_{id}, y_{id})\sim \mathcal{S}_{ID}$ and $(\mathbf{x}_{ood}, y_{ood})\sim\mathcal{S}_{OOD}$ represent two data points sampled from the ID data $\mathcal{S}_{ID}$ and OOD data $\mathcal{S}_{OOD}$ respectively, and, define $\mathbf{p}_{id} = \underset{y_{id}}{\max}\,p(y_{id}\vert f_{\theta}(\mathbf{x}_{id}))$ to be the maximum softmax probability (MSP) for $\mathbf{x}_{id}\in\mathcal{S}_{ID}$, and, $\mathbf{p}_{ood} = \underset{y_{id}}{\max}\, p(y_{id}\vert f_{\theta}(\mathbf{x}_{ood}))$ be the MSP for $\mathbf{x}_{ood}\in\mathcal{S}_{OOD}$. Then our objective is formulated as:
\begin{align}
    L(\mathbf{x}_{id}, \mathbf{x}_{ood}, y_{id})
    &= \underbrace{-\mathbb{E}\left[\log\,p(y_{id}\vert\mathbf{x}_{id})\right]}_{\text{cross-entropy}}\;\;+\underbrace{\lambda \, \operatorname{cos}(\mathbf{p}_{id}, \mathbf{p}_{ood})}_{\text{cosine-regularisation}}
\end{align}

The regularisation strength $\lambda$ is often obtained using the validation set, and whenever $\lambda = -1$ then the underlying objective becomes a minimax optimisation formulation similar to adversarial learning paradigms~\cite{Pang2020Rethinking}, with the advantage that it is faster to train a model with this objective since it avoids computing gradients for worst-case perturbations on the inputs. The goal is to lower the cross-entropy error on $\mathcal{S}_{ID}$ while at the same time increasing the cosine angle between $\mathcal{S}_{ID}$ and $\mathcal{S}_{OOD}$. This synergy of minimax optimisation can also be found in energy-based models~\cite{liu2020,grathwohl2019your} where the intention is to lower the energy for similar samples while at the same time increasing the energy on dissimilar inputs. This approach is summarised in Algorithm \ref{alg:alg1}.

\begin{algorithm}[H]
\begin{algorithmic}
    \Procedure{ContReg}{$\mathbf{x}_{id}, \mathbf{x}_{ood}, y_{id}$} 
    \State $f_{\theta} \gets  \theta$ \Comment{initialise model}\;
    \State $\mathbf{z}_{id}, \mathbf{z}_{ood} \gets f_{\theta}(\mathbf{x}_{id}, \mathbf{x}_{ood})$ \Comment{compute logits for $\mathbf{x}_{id}\in\mathcal{S}_{ID}, \mathbf{x}_{ood}\in\mathcal{S}_{OOD}$}\;
    \State $\widehat{\mathbf{p}}_{id}, \widehat{\mathbf{p}}_{ood}  \gets softmax(\mathbf{z}_{id}, \mathbf{z}_{ood})$ \Comment{probab. for logits $\in(\mathcal{S}_{ID}, \mathcal{S}_{OOD})$}
    \State $CE \gets -\mathbb{E}[\log\;p(y_{id}\vert\mathbf{x}_{id})]$ \Comment{compute cross-entropy for $(\mathbf{x}_{id}, y_{id})\in\mathcal{S}_{ID}$}\;
    \State $\operatorname{cos} \gets \frac{\mathbf{p}_{id}\top\mathbf{p}_{ood}}{\Vert\mathbf{p}_{id}\Vert\,\Vert\mathbf{p}_{ood}\Vert}$ \Comment{computer cosine for probabilities $\widehat{\mathbf{p}}_{id}, \widehat{\mathbf{p}}_{ood}$} \;
    \State $L \gets CE + \lambda\, \operatorname{cos}$ \Comment{compute final regularised loss}
    \State $\theta_{t+1} = \theta_{t} - \eta\,\nabla_{\theta}L$ \Comment{compute gradient w.r.t params $\theta$ and backprop errors}
    \EndProcedure
    \caption{Contrastive Regularised Objective}
\label{alg:alg1}
\end{algorithmic}
\end{algorithm}

The second contribution is a novel ranking objective for OOD detection. We utilise cosine similarity as a metric learning function in addition to explicit $\ell^{2}$ and $\ell^{1}$ regularisation for $\mathcal{S}_{ID}$ and $\mathcal{S}_{OOD}$ respectively, which essentially constitutes into the following objective:
\begin{align}
    L(\mathbf{x}_{id}, \mathbf{x}_{ood}, y_{id}) 
    &= \underbrace{\max(0, \gamma + \operatorname{cos}(\mathbf{p}_{id}, \mathbf{p}_{ood}))}_{\text{ranking objective}}
    & + \underbrace{\lambda_{1}\,\sum_{n}\vert\mathbf{p}_{ood} - 1/k\vert}_{\ell_{1}\text{-regularisation on }\mathcal{S}_{OOD}}\nonumber \\
    & + \underbrace{\lambda_{2}\,\sum_{n}\Vert y_{id}\,\mathbf{p}_{id} - \alpha\Vert_{2}}_{\ell_{2}\text{-regularisation on }\mathcal{S}_{ID}}
\end{align}

Notice that $k\in\mathbb{Z}$ refers to the number of ID classes in $\mathcal{S}_{ID}$, and $y_{id}$ represents a one-hot encoding of the labels, while $\alpha\in\mathbb{R}$ is a user defined scalar that indicates the desired ID accuracy. There are a number of hyperpameters $\left\{\gamma, \lambda_{1}, \lambda_{2}\right\}$ which can be tuned on the validation set, $\gamma$ defines the margin and $\lambda_{1},\,\lambda_{2}$ refer to the regularisation strength. This approach is depicted in Algorithm \ref{alg:alg2}.

\begin{algorithm}[H]
\begin{algorithmic}
    \Procedure{ContRank}{$\mathbf{x}_{id}, \mathbf{x}_{ood}, y_{id}$} 
    \State $f_{\theta} \gets  \theta$ \Comment{initialise model}\;
    \State $\mathbf{z}_{id}, \mathbf{z}_{ood} \gets f_{\theta}(\mathbf{x}_{id}, \mathbf{x}_{ood})$ \Comment{compute logits for $\mathbf{x}_{id}\in\mathcal{S}_{ID}, \mathbf{x}_{ood}\in\mathcal{S}_{OOD}$}\;
    \State $\widehat{\mathbf{p}}_{id}, \widehat{\mathbf{p}}_{ood}  \gets softmax(\mathbf{z}_{id}, \mathbf{z}_{ood})$ \Comment{probab. for logits $\in(\mathcal{S}_{ID}, \mathcal{S}_{ood})$}
    \State $\ell_{1} \gets \lambda_{1}\,\sum_{n}\vert\mathbf{p}_{ood} - 1/k\vert$ \Comment{compute $\ell_{1}$-regularisation for $\widehat{\mathbf{p}}_{ood}\in\mathcal{S}_{ood}$}\;
    \State $\ell_{2} \gets \lambda_{2}\,\sum_{n}\Vert y_{id}\,\mathbf{p}_{id}, \alpha\Vert$ \Comment{compute $\ell_{2}$-regularisation for $\widehat{\mathbf{p}}_{id}\in\mathcal{S}_{id}$}\;
    \State $\operatorname{cos} \gets \frac{\mathbf{p}_{id}\top\mathbf{p}_{ood}}{\Vert\mathbf{p}_{id}\Vert\,\Vert\mathbf{p}_{ood}\Vert}$ \Comment{compute cosine for probabilities $\widehat{\mathbf{p}}_{id}, \widehat{\mathbf{p}}_{ood}$} \;
    \State $L \gets \max(0, \gamma + \operatorname{cos}(\cdot)) + \ell_{1} + \ell_{2}$ \Comment{compute the final ranking loss}
    \State $\theta_{t+1} = \theta_{t} - \eta\,\nabla_{\theta}L$ \Comment{compute gradient w.r.t params $\theta$ and backprop errors}
    \EndProcedure
    \caption{Contrastive Ranking Objective}
\label{alg:alg2}
\end{algorithmic}
\end{algorithm}

\section{Experiments}

In this section we describe a set of experiments designed to evaluate the effectiveness of the proposed objectives defined in the previous section, and to compare them to existing approaches. We first evaluate the objectives using an artificially generated dataset, before using a selection of real image classification datasets to evaluate them.

\subsection{Artificial Data Experiments}
\begin{figure}[ht!]
    \centering
    \includegraphics[width=\textwidth]{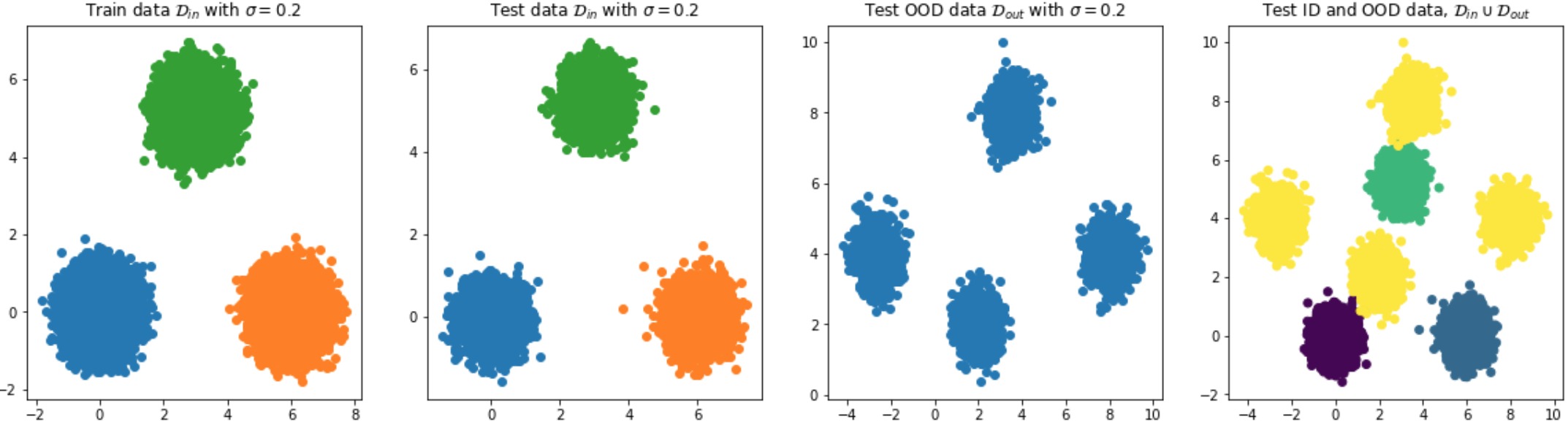}
    \caption{Synthetic dataset from left to right, comprised of ID train data (1st), ID test data (2nd), OOD test data (3rd), and finally the union of $\text{ID} \cup \text{OOD}$ (yellow) test data (4th).}
\label{fig:synthetic_dataset}
\end{figure}

To validate the efficacy of our proposed objectives for OOD detection we designed a controlled experiment utilising synthetic data. The training ID data $\mathcal{S}_{ID}$ is comprised of 3 Gaussians with standard deviation $\sigma$ representing different classes in a multi-class classification setting. The different subset splits \textit{train} $\sim\mathcal{S}_{ID}^{train}$, \textit{test} $\sim\mathcal{S}_{ID}^{test}$ and \textit{test OOD} $\sim\mathcal{S}_{OOD}$ over the synthetic dataset are depicted in Figure~\ref{fig:synthetic_dataset}.

The underlying model is a 3-layer MLP network, trained on the synthetic ID train split. During inference we test each objective on the OOD test data $\mathcal{S}_{OOD}$ constructed of 4-Gaussians displaced at different locations. To measure performance at OOD detection we measure AUC based on three OOD metrics on the models' logits: \textit{confidence}, \textit{entropy}, and \textit{mutual information}.

\subsubsection{Results \& Discussion} 
\label{sec:ranking_objectives_artificial_data}

\begin{figure}[ht!]
    \centering
    \begin{subfigure}[t]{\textwidth}
        \includegraphics[width=\textwidth]{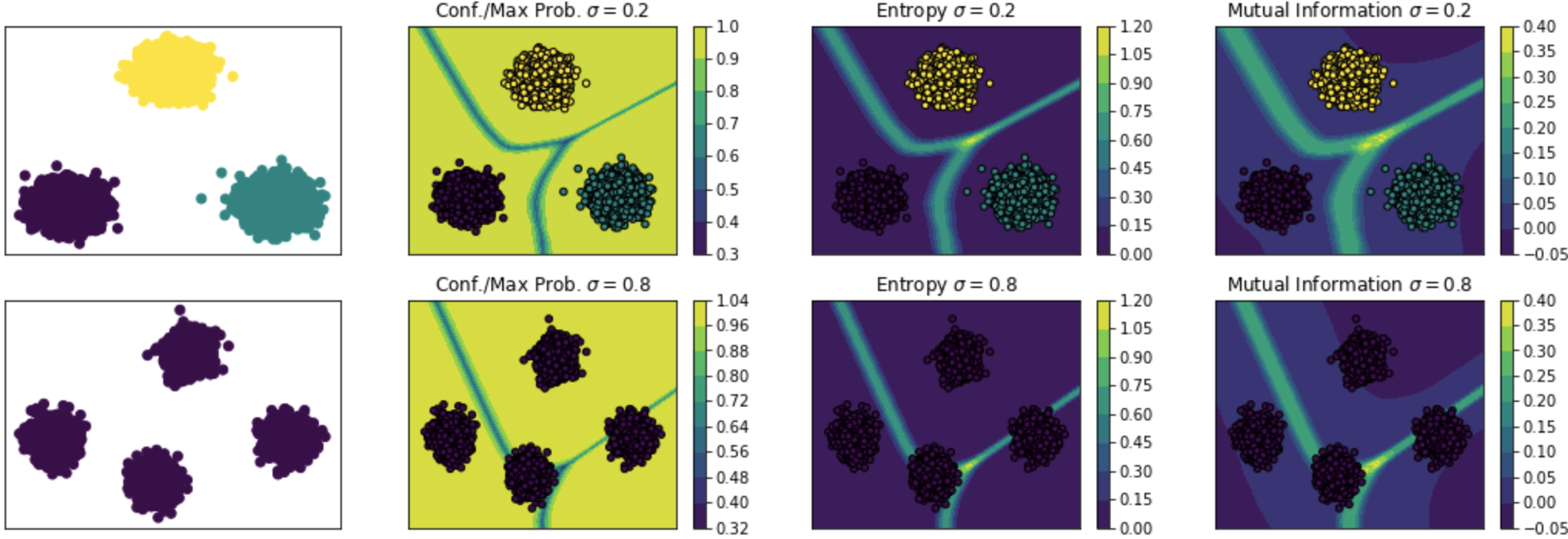}
        \caption{CrossEntropy objective as baseline.}
    \end{subfigure}
    
    \begin{subfigure}[t]{\textwidth}
        \includegraphics[width=\textwidth]{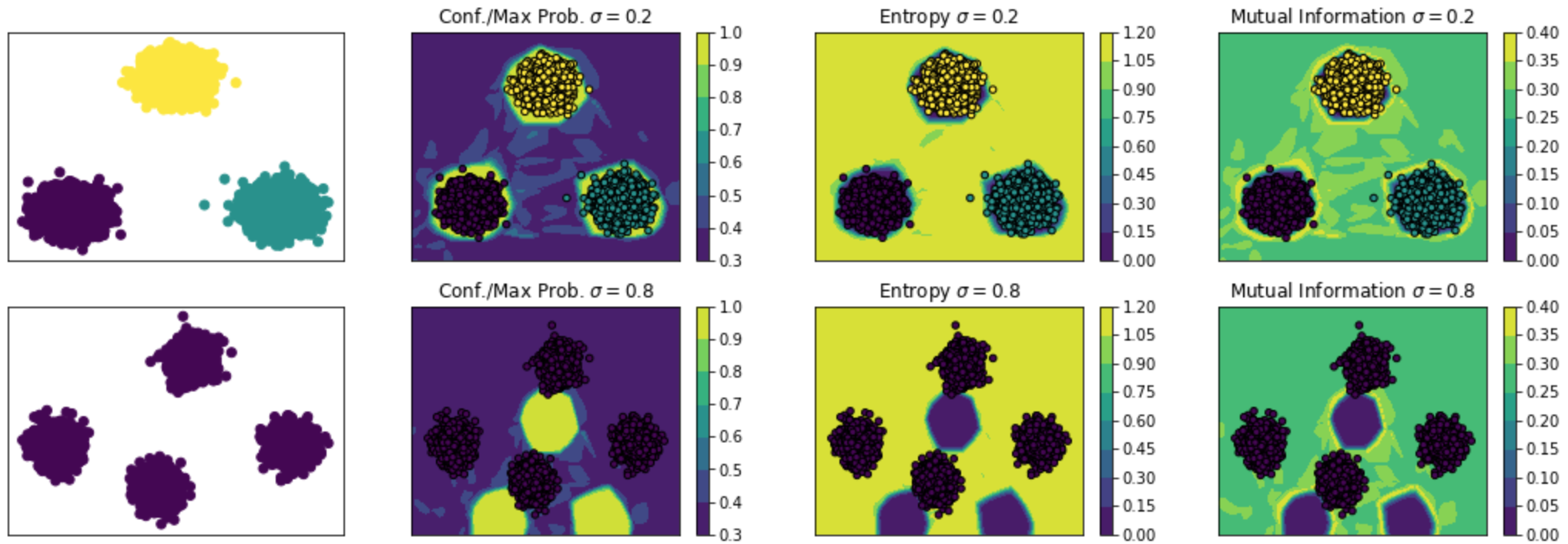}
        \caption{ContReg (ours)}
    \end{subfigure}
    
    \begin{subfigure}[t]{\textwidth}
        \includegraphics[width=\textwidth]{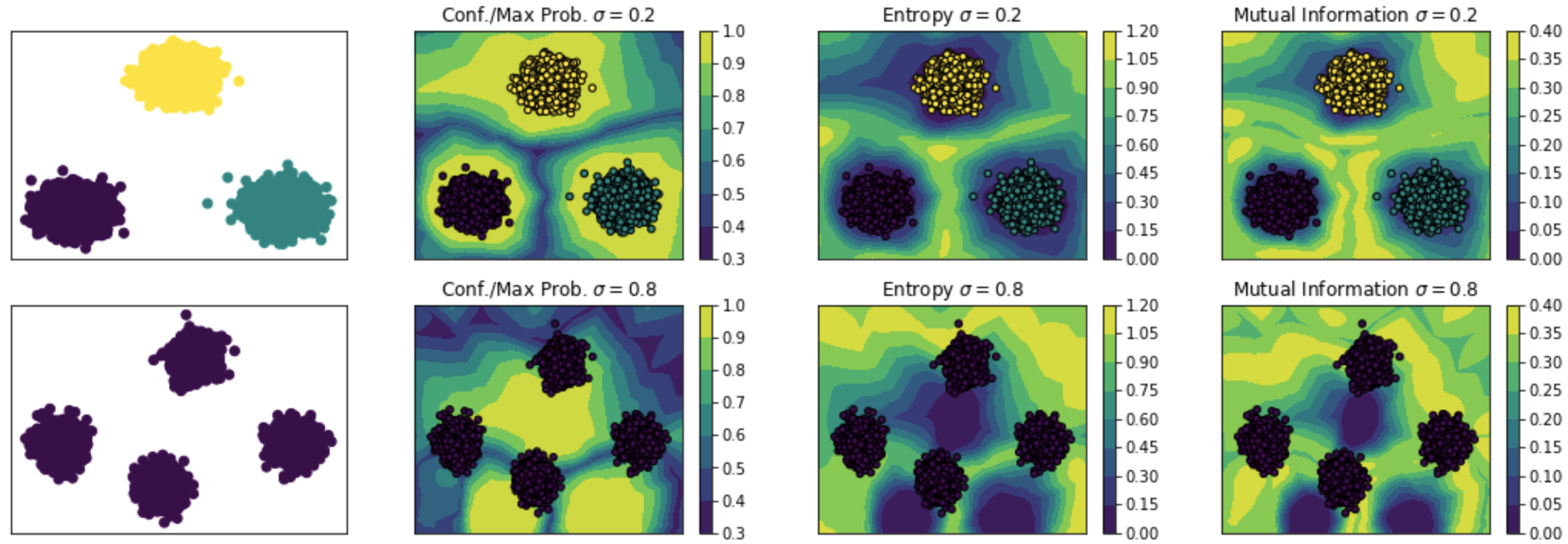}
        \caption{ContRank (ours)}
    \end{subfigure}
\caption{Decision boundaries across objective functions for ID (1st row) and OOD (2nd row) test data.}
\label{fig:synthetic_data_decision_boundaries}
\end{figure}

Table~\ref{tab:synthetic_data_auroc} presents our findings for OOD detection across objectives, while Figure~\ref{fig:synthetic_data_decision_boundaries} depicts the different decision boundaries across each objective for ID (1st row) and OOD (2nd row) test data.
\begin{table}[ht!]
\caption{Accuracy and AUC-ROC-scores across objectives \& metrics represented in percentage (\%).}
\centering
\resizebox{\textwidth}{!}{%
\begin{tabular}{cc|l|l|lll} 
\hline
\multicolumn{2}{c|}{Data} & \multirow{2}{*}{Objectives} & \multirow{2}{*}{Accuracy} & \multicolumn{3}{c}{AUC-ROC scores} \\ 
\cline{1-2}\cline{5-7}
$\mathcal{S}_{ID}$ & $\mathcal{S}_{OOD}$ &  &  & Confidence & Entropy & Mutual Information \\ 
\hline
3-Gaussians & 4-Gaussians & CrossEntropy & 100 & 61.64 & 61.61 & 63.62 \\
 &  & CrossEntropy+MC-Dropout & 100 & 75.14 & 73.63 & 73.56 \\
 &  & ContReg (ours) & 100 & 99.99 & 99.99 & 99.99 \\
 &  & ContRank (ours) & 100 & 99.99 & 99.99 & 99.99
\end{tabular}
}
\label{tab:synthetic_data_auroc}
\end{table}

As suggested in Table~\ref{tab:synthetic_data_auroc} our proposed methods achieve near optimal OOD detection when presented with ambiguous test data. Notice that explicit regularisation (e.g. MC-Dropout) does indeed provide additional benefit in OOD detection. Similar conclusions supporting our claims have been demonstrated in prior works of~\cite{sagawa2020,rice2020,khani2020,wei2020,khani2021} regarding the impact of regularisation. To under stand why explicit regularisation improves OOD detection we exhibit the existence of a connection among Dropout~\cite{gal2016}, Mixup~\cite{zhang2017} and Randomised smoothing~\cite{cohen2019}, where these methods act as boundary thickness~\cite{yang2020}. It is evident from Figure~\ref{fig:synthetic_data_decision_boundaries} that a model trained with CrossEntropy on a classification setting acts as a max-margin predictor while our objective act as density estimator. This indicates that the choice of objective and regularisation play a crucial role in the final behaviour of the predictor.

\subsection{Real Data Experiment}
Five well-known image classification datasets are used in this experiment:\textit{CIFAR-10}, \textit{CIFAR-100}, \textit{SVHN}, \textit{FashionMNIST} and \textit{LSUN}. Every dataset was split into three distinct sets \{\textit{train, validation, test}\} with random mirroring and cropping  augmentations. We utilised WideResNet28x10~\cite{zagoruyko2016wide} as the DNN model trained for 300 epochs using a validation set for hyper-parameter tuning and rolling back to the best checkpoint to avoid overfitting. The optimiser was Stochastic Gradient Descent (SGD)~\cite{robbins1951,kiefer1952} with momentum set to 0.9 and weight decay in the range $[3e^{-4}, 5e^{-4}]$. Given that all datasets have balanced class distributions we utilised classification accuracy to measure their performance on the clean ID test data. To measure the ability of the models to recognise OOD examples we utilised the predictions of the test portion of three OOD datasets (see Table~\ref{tab:real_data_aucroc}). We measure the separation between ID and OOD data using the area under the curve (AUC-ROC) for each approach.

We also compare the effectiveness of the custom objective described in this paper with the following existing objectives designed to address the OOD problem.

\begin{tabular}{lp{0.8\textwidth}}
Mahalanobis~\cite{lee2018} &
\(
M(\mathbf{x})=\max _{c}-(f(\mathbf{x})-\widehat{\mu}_{c})^{\top} \widehat{\mathbf{\Sigma}}^{-1}\left(f(\mathbf{x})-\widehat{\mu}_{c}\right)
\) \\ \\

ODIN~\cite{liang2018enhancing} & 
\(
g(\boldsymbol{x} ; \delta, T, \varepsilon)=
\begin{cases}
    1 & \text{if } \max_{i} p(\tilde{\mathbf{x}} ; T) \leq \delta \\ 
    0 & \text{if }\max_{i} p(\tilde{\mathbf{x}} ; T) > \delta
\end{cases}
\) \\ \\

MSRep~\cite{shalev2018} & 
\(
\bar{\ell}(\mathbf{x}, y ; \theta)=\sum_{k=1}^{K} d_{\cos }\left(e^{k}(y), f_{\theta^{k}}^{k}(\mathbf{x})\right)
\) \\ \\

OutlierExposure~\cite{hendrycks2018_outlier_exposure} & 
\(
\mathbb{E}_{(x, y) \sim \mathcal{D}_{\text {in }}}\left[\mathcal{L}(f(x), y)+\lambda \mathbb{E}_{x^{\prime} \sim \mathcal{D}_{\text {out }}^{\text {OE }}}\left[\mathcal{L}_{\mathrm{OE}}\left(f\left(x^{\prime}\right), f(x), y\right)\right]\right]
\) \\ \\

EnergyOOD~\cite{liu2020} & 
\(
\min_{\theta} \mathbb{E}_{(\mathbf{x}, y) \sim \mathcal{D}_{\mathrm{in}}}\left[-\log F_{y}(\mathbf{x})\right]+\lambda \cdot L_{\mathrm{energy}}
\) \\ \\

CSI~\cite{jihoon2020} & 
\(
-\frac{1}{\left|\left\{x_{+}\right\}\right|} \log \frac{\sum_{x^{\prime} \in\left\{x_{+}\right\}} \exp \left(\operatorname{sim}\left(z(x), z\left(x^{\prime}\right)\right) / \tau\right)}{\sum_{x^{\prime} \in\left\{x_{+}\right\} \cup\left\{x_{-}\right\}} \exp \left(\operatorname{sim}\left(z(x), z\left(x^{\prime}\right)\right) / \tau\right)}
\) \\ \\

DoSE~\cite{morningstar2021} & 
\(
\frac{1}{m} \sum_{j}^{m}(\log q(x_{j}\vert\left\{x_{i}\right\}_{i}^{n}, T, \gamma)^{2}
+2 \mathbb{H}[p] \frac{1}{m} \sum_{j}^{m} \log q(x_{j} \vert\left\{x_{i}\right\}_{i}^{n}, T, \gamma)
\)
\end{tabular}

\subsubsection{Results \& Discussion}
\label{sec:ranking_objectives_real_data}
\begin{figure}[ht!]
    \centering
    \includegraphics[width=\textwidth]{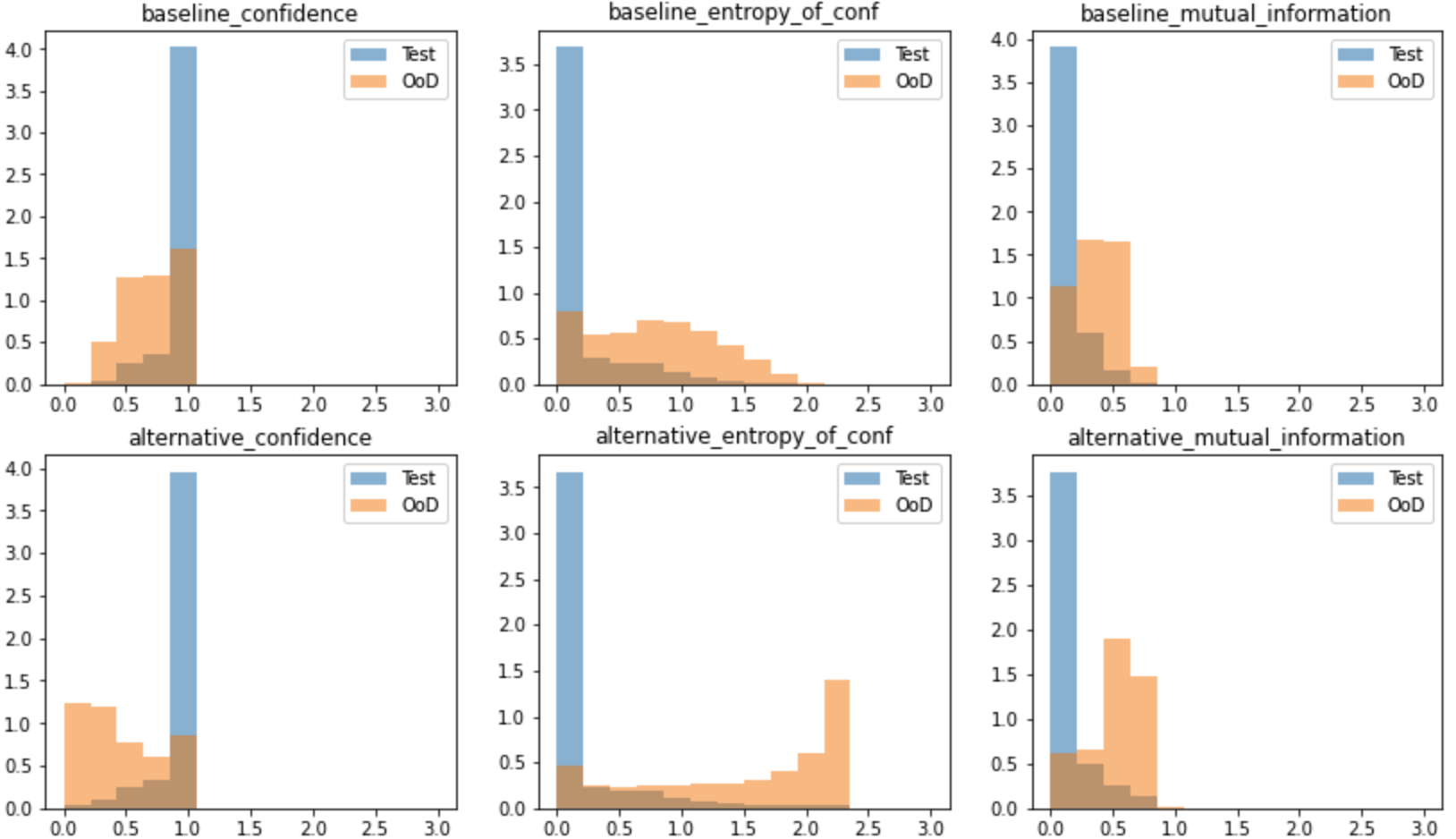}
    \caption{Comparison of CrossEntropy (baseline) against our proposed objective (i.e. ContReg (see Table~\ref{tab:related_work_comparison} and \ref{tab:real_data_aucroc}) across three metrics \{\textit{confidence, entropy, mutual information}\} with respect to a WideResNet28x10 architecture.}
    \label{fig:histogram_objectives}
\end{figure}

According to Table~\ref{tab:related_work_comparison} and Table~\ref{tab:real_data_aucroc} our objectives outperform the max softmax probabilty (MSP) baseline by a large margin (see Figure~\ref{fig:histogram_objectives}), except when $\{\textit{CIFAR-100}\}\in\mathcal{S}_{ID}$ while $\{\textit{CIFAR-10, LSUN}\}\in\mathcal{S}_{OOD}$. This observation is interesting since it suggests that the value of auxiliary information from $\mathcal{S}_{OOD}$ might be degrading when $\mathcal{S}_{ID}\supseteq\mathcal{S}_{OOD}$. With the term superset $\mathcal{S}_{ID}\supseteq\mathcal{S}_{OOD}$ we refer to the fact that the ID data $\mathcal{S}_{ID}$ might represent a broader set of features compared to OOD data $\mathcal{S}_{OOD}$. Thus, training a model with a small subset of the ID data as OOD might not be beneficial since no additional information is presented to the model because the features from $\mathcal{S}_{ID}$ and $\mathcal{S}_{OOD}$ conflict with each other. Another factor that impairs OOD detection is the presence of label noise~\cite{mitros2020}, (e.g. $\{\textit{CIFAR-10 vs. CIFAR-100}\}$), which has been identified with the term \textit{near-OOD vs. far-OOD} in subsequent work~\cite{winkens2020}. A natural question arising from this observation is whether we can identify the inflection point between label noise and OOD detection?
\begin{table}[ht!]
\centering
\caption{Accuracy of models on ID dataset classification tasks.}
\label{tab:real_data_accuracy}
\begin{tabular}{l|llll} 
\hline
Model                      & CIFAR-10 & \multicolumn{1}{c}{SVHN} & \multicolumn{1}{c}{FashionMNIST} & \multicolumn{1}{c}{CIFAR100}  \\ 
\hline
DNN                        & 95.06    & 96.67   & 95.27   & 77.44  \\
DPN                        & 88.10    & 90.10   & 93.20   & 79.34  \\
MC-Dropout                 & 96.22    & 96.90   & 95.40   & 78.39  \\
SWAG                       & 96.53    & 97.06   & 93.80   & 78.61  \\
JEM                        & 92.83    & 96.13   & 83.21   & 77.86  \\
CE+$\ell_{1}$              & 90.66    & 95.34   & 93.89   & 62.30  \\
CE+$\ell_{1}$+MCD          & 90.33    & 94.85   & 91.37   & 60.35  \\
ContReg (ours)             & 90.76    & 95.25   & 93.68   & 72.78  \\
ContReg+MCD                & 90.31    & 94.75   & 93.01   & 64.04  \\
ContRank (ours)            & 89.01    & 94.97  & 93.40    & 64.32  \\
ContRank+MCD               & 91.96    & 82.34   & 93.13  & 60.43                        
\end{tabular}
\end{table}

Table~\ref{tab:related_work_comparison} compares the performance of the proposed objectives described in this paper with existing OOD detection approaches, where we report confidence-based AUC-ROC scores matching our experimental setting. Our methods outperform \textit{MSP}, \textit{ODIN}, and \textit{EnergyOOD} (w/o pre-training), and provide comparable results with \textit{Mahalanobis}, \textit{MSRep}, and \textit{OE}. 
\begin{table}[ht!]
\centering
\caption{Comparison of our methods with related work based on published results in the literature corresponding with our setting.}
\label{tab:related_work_comparison}
\resizebox{\linewidth}{!}{%
\begin{tabular}{ll|llllllll} 
\hline
\multicolumn{2}{c|}{Data} & \multicolumn{8}{c}{\multirow{2}{*}{AUC-ROC scores}} \\ 
\cline{1-2}
\multirow{2}{*}{$\mathcal{S}_{ID}$} & \multirow{2}{*}{$\mathcal{S}_{OOD}$} & \multicolumn{8}{l}{} \\ 
\cline{3-10}
 &  & MSP & Mahalanobis & ODIN & MSRep & OE & EnergyOOD & ContReg(ours) & ContRank(ours) \\ 
\hline
CIFAR-10 & CIFAR-100* & 86.15 & 93.90 & 85.59 & 91.23 & 93.30 & 92.60 & 92.23 & 94.23 \\
 & SVHN & 89.60 & 97.62 & 91.96 & 99.48 & 98.40 & 90.96 & 99.18 & 95.40 \\
 & LSUN & 88.54 & 96.30 & 90.35 & 96.05 & 97.60 & 94.24 & 92.44 & 94.77 \\
 &  &  &  &  &  &  &  &  &  \\
CIFAR-100 & CIFAR-10 & 73.41 & 81.34 & 74.54 & 81.49 & 75.70 & 76.61 & 72.94 & 68.89 \\
 & SVHN* & 71.44 & 86.01 & 67.26 & 87.42 & 86.66 & 73.99 & 99.68 & 99.95 \\
 & LSUN & 75.38 & 93.9 & 78.94 & 79.05 & 79.71 & 79.23 & 70.50 & 62.17
\end{tabular}
}
\end{table}

From Table~\ref{tab:real_data_aucroc} we can observe that even though explicit regularisation overall is beneficial compared to no regularisation, on the contrary, stronger regularisation might deteriorate OOD detection. An ongoing inquiry is to formally characterise and identify the necessary and sufficient conditions of regularisation in order to robustify models against ambiguous and corrupted inputs.

To evaluate whether our method is robust against common corruptions we utilised CIFAR10-C and CIFAR100-C. Similar to~\cite{hendrycks2019} we report the mean corruption error (mCE) in Table~\ref{tab:corruptions}, with the exception that we do not adjust for the varying corruption difficulties by dividing the average corruption error with those of a baseline model. Observe that our objective attains the smallest mCE on CIFAR10-C indicating that is indeed robust against common corruptions while on CIFAR100-C cross-entropy with $\ell_{1}$-regularisation attain the smallest error with ours being second best.
\begin{figure}[ht!]
    \centering
    \begin{subfigure}[t]{0.45\textwidth}
        \centering
        \includegraphics[width=\textwidth]{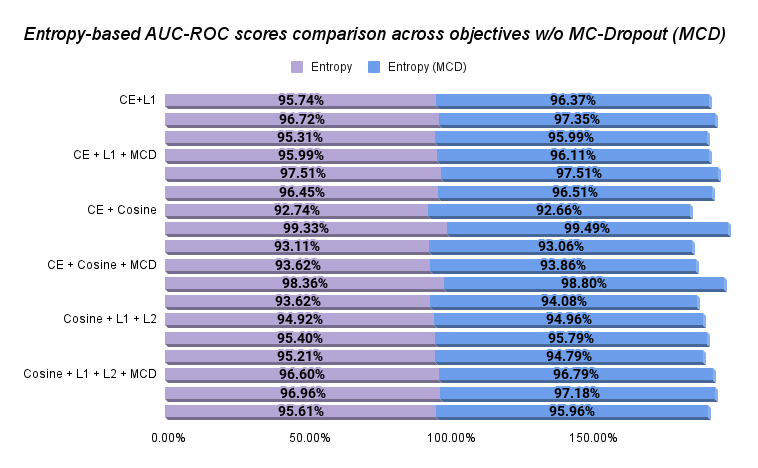}
    \caption{Train, \{\textit{CIFAR-10}\}, Test \{\textit{CIFAR-100, SVHN, LSUN}\}}
    \end{subfigure}
    ~
    \begin{subfigure}[t]{0.45\textwidth}
        \centering
        \includegraphics[width=\textwidth]{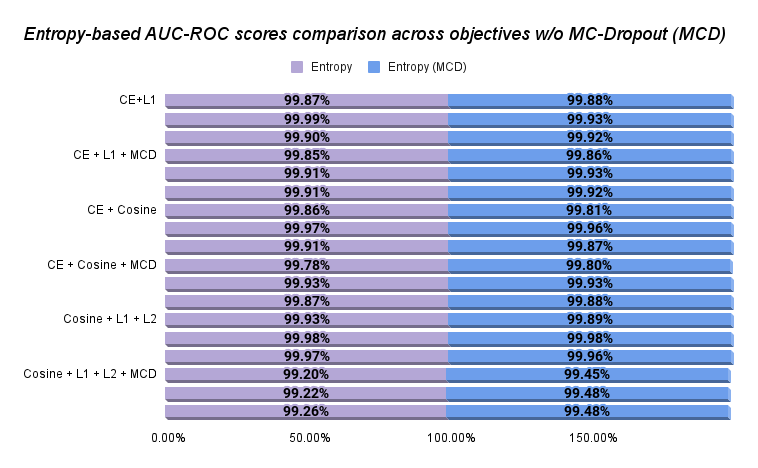}
    \caption{Train, \{\textit{SVHN}\}, Test \{\textit{CIFAR-100, CIFAR-10, LSUN}\}}
    \end{subfigure}
    
     \begin{subfigure}[t]{0.45\textwidth}
        \centering
        \includegraphics[width=\textwidth]{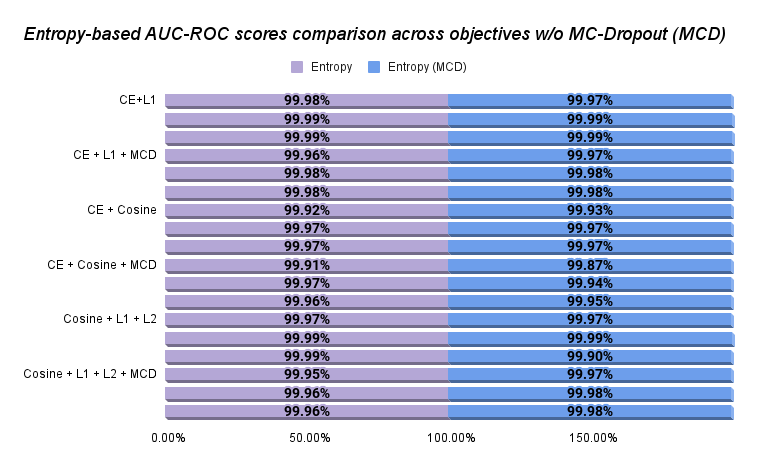}
    \caption{Train, \{\textit{FashionMNIST}\}, Test \{\textit{CIFAR-100, CIFAR-10, LSUN}\}}
    \end{subfigure}
    ~
    \begin{subfigure}[t]{0.45\textwidth}
        \centering
        \includegraphics[width=\textwidth]{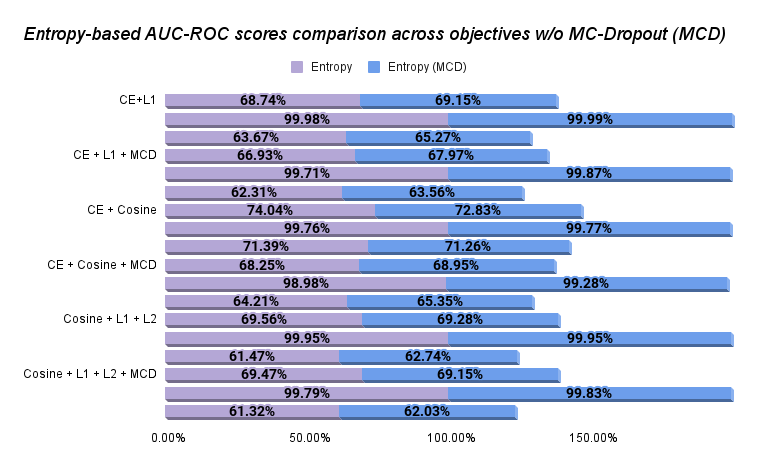}
    \caption{Train, \{\textit{CIFAR-100}\}, Test \{\textit{CIFAR-10, SVHN, LSUN}\}}
    \end{subfigure}
\caption{Comparison of objective functions with and without MC-Dropout during inference. Each histogram corresponds to evaluating a particular loss on a different OOD dataset.}
\label{fig:mcd_objectives_comparison}
\end{figure}

\begin{table}[ht!]
\centering
\caption{Evaluating objective functions across common corruptions against CIFAR10-C and CIFAR100-C measured in average corruption error (mCE).}
\resizebox{0.8\textwidth}{!}{%
\begin{tabular}{l|cc} 
\hline
\multirow{2}{*}{Objectives} & \multicolumn{2}{c}{mCE} \\ 
\cline{2-3}
& CIFAR10-C & CIFAR100-C \\ 
\hline
CrossEntropy & 161.14 & 717.04 \\
CrossEntropy+MCD & 120.91 & 536.63 \\
CrossEntropy+$\ell_{1}$ & 144.02 & 247.78 \\
CrossEntropy+$\ell_{1}$+MCD & 140.96 & 285.04 \\
ContReg (ours) & 119.98 & 337.94 \\
ContReg+MCD & 129.20 & 269.52  \\
ContRank (ours) & 149.46 & 258.73  \\
ContRank+MCD & 167.30 & 306.42
\end{tabular}
}
\label{tab:corruptions}
\end{table}

\begin{sidewaystable}[ht!]
\centering
\caption{Out-of-distribution results.  AUC-ROC scores based on entropy. The values in bracket are \% improvement of algorithm w.r.t. baseline. An $\uparrow$ indicates improvement \&  $\downarrow$ degradation. The asterisk (*) indicates auxiliary information during training. \label{tab:ood_full_tab}}
\resizebox{\columnwidth}{!}{%
\begin{tabular}{ll|lllllllllll} 
\hline
\multicolumn{2}{c|}{Data}                             & \multicolumn{1}{l|}{(baseline)} & \multicolumn{10}{c}{Entropy AUC-ROC score (\% gain wrt. baseline)}                                                                                                                                                                                                                                               \\ 
\cline{3-13}
$\mathcal{S}_{ID}$  &  $\mathcal{S}_{OOD}$                    & DNN                             & DPN                                    & MCD                             & SWAG                                   & JEM                                      & CE+$\ell_{1}$          & CE+$\ell_{1}$+MCD & ContReg(ours)     & ContReg+MCD & ContRank(ours) & ContRank+MCD  \\ 
\hline
\multirow{3}{*}{CIFAR-10}  & CIFAR-100*  & 86.27  & 85.60 ($\downarrow$0.78\%)  & 89.92 ($\uparrow$4.23\%)  & 91.89 ($\uparrow$6.51\%)  & 87.35 ($\uparrow$1.25\%)  & 95.74 ($\uparrow$10.97\%)  & 95.99 ($\uparrow$11.26\%)  & 92.74 ($\uparrow$7.50\%)  & 93.62 ($\uparrow$8.52\%) & 94.92 ($\uparrow$10.03\%)  & 96.60 ($\uparrow$11.97\%)    \\
                              & SVHN                  & 89.72                           & 98.90 ($\uparrow$10.23\%)              & 96.25 ($\uparrow$7.28\%)               & 98.62 ($\uparrow$9.92\%)               & 89.22 ($\downarrow$0.56\%)               & 96.72 ($\uparrow$7.80\%)   & 97.51 ($\uparrow$8.68\%)    & 99.33 ($\uparrow$10.71\%) & 98.36 ($\uparrow$9.63\%)  & 95.40 ($\uparrow$6.33\%)  & 96.96 ($\uparrow$8.07\%)   \\
                              & LSUN                  & 88.83                           & 83.30 ($\downarrow$6.23\%)             & 92.04 ($\uparrow$3.61\%)               & 95.12 ($\uparrow$7.08\%)               & 89.84 ($\uparrow$1.14\%)                 & 95.31 ($\uparrow$7.29\%)   & 96.45 ($\uparrow$8.57\%)    & 93.11 ($\uparrow$4.82\%)  & 93.62 ($\uparrow$5.39\%)        & 95.21 ($\uparrow$7.18\%)        & 95.61 ($\uparrow$7.63\%)   \\
\multicolumn{1}{c}{}          & \multicolumn{1}{c|}{} &                                 &                                        &                                        &                                        &                                          &                  &                   &                 &                       &                       &     \\
\multirow{3}{*}{SVHN}         & CIFAR-100             & 93.19                           & 99.10 ($\uparrow$6.34\%)               & 94.33 ($\uparrow$1.22\%)               & 95.97 ($\uparrow$2.98\%)               & 92.34 ($\downarrow$0.91\%)               & 99.87 ($\uparrow$7.17\%)   & 99.85 ($\uparrow$7.15\%)    & 99.86 ($\uparrow$7.16\%)  & 99.78 ($\uparrow$7.07\%)        & 99.93 ($\uparrow$7.23\%)        & 99.20 ($\uparrow$6.45\%)      \\
                              & CIFAR-10*             & 94.58                           & 99.60 ($\uparrow$5.31\%)               & 94.97 ($\uparrow$0.41\%)               & 96.03 ($\uparrow$1.53\%)               & 92.85 ($\downarrow$1.83\%)               & 99.99 ($\uparrow$5.72\%)   & 99.91 ($\uparrow$5.64\%)    & 99.97 ($\uparrow$5.70\%)  & 99.93 ($\uparrow$5.66\%)        & 99.98 ($\uparrow$5.71\%)        & 99.22 ($\uparrow$4.91\%)      \\
                              & LSUN                  & 92.97                           & 99.70 ($\uparrow$7.24\%)               & 93.31 ($\uparrow$0.37\%)               & 95.71 ($\uparrow$2.95\%)               & 91.82 ($\downarrow$1.24\%)               & 99.90 ($\uparrow$7.45\%)   & 99.91 ($\uparrow$7.46\%)    & 99.91 ($\uparrow$7.46\%)  & 99.87 ($\uparrow$7.42\%)        & 99.97 ($\uparrow$7.53\%)         & 99.26 ($\uparrow$6.77\%)  \\
\multicolumn{1}{c}{}          & \multicolumn{1}{c|}{} &                                 &                                        &                                        &                                        &                                          &                  &                   &                 &                       &                       &     \\
\multirow{3}{*}{FashionMNIST} & CIFAR-100             & 91.20                           & 99.50 ($\uparrow$9.10\%)               & 93.75 ($\uparrow$2.80\%)               & 96.19 ($\uparrow$5.47\%)               & 62.79 ($\downarrow$31.15\%)              & 99.98 ($\uparrow$9.63\%)   & 99.96 ($\uparrow$9.61\%)    & 99.92 ($\uparrow$9.56\%)  & 99.91 ($\uparrow$9.55\%)        & 99.97 ($\uparrow$9.62\%)        & 99.95 ($\uparrow$9.59\%)     \\
                              & CIFAR-10*             & 94.59                           & 99.60 ($\uparrow$5.30\%)               & 96.06 ($\uparrow$1.55\%)               & 94.28 ($\downarrow$0.33\%)             & 64.76 ($\downarrow$31.54\%)              & 99.99 ($\uparrow$5.71\%)   & 99.98 ($\uparrow$5.70)\%    & 99.97 ($\uparrow$5.69\%)  & 99.97 ($\uparrow$5.69\%)        & 99.99 ($\uparrow$5.71\%)        & 99.96 ($\uparrow$5.68\%)      \\
                              & LSUN                  & 93.34                           & 99.80 ($\uparrow$6.92\%)               & 97.40 ($\uparrow$4.35\%)               & 99.05 ($\uparrow$6.12\%)               & 65.38 ($\downarrow$29.96\%)              & 99.99 ($\uparrow$7.12\%)   & 99.98 ($\uparrow$7.11)\%    & 99.97 ($\uparrow$7.10\%)  & 99.96 ($\uparrow$7.09\%)        & 99.99 ($\uparrow$7.12\%)        & 99.96 ($\uparrow$7.09\%)   \\
\multicolumn{1}{c}{}          & \multicolumn{1}{c|}{} &                                 &                                        &                                        &                                        &                                          &                  &                   &                 &                       &                       &      \\
\multirow{3}{*}{CIFAR-100}    & CIFAR-10              & 78.25                           & 85.15 ($\uparrow$8.82\%)               & 80.70 ($\uparrow$3.13\%)               & 84.92 ($\uparrow$8.52\%)               & 77.64 ($\downarrow$0.78\%)               & 68.74 ($\downarrow$12.15\%) & 66.93 ($\downarrow$14.46\%)  & 74.04 ($\downarrow$5.38\%) & 68.25 ($\downarrow$12.77\%)      & 69.56 ($\downarrow$11.10\%)      & 69.47 ($\downarrow$11.22\%)   \\
                              & SVHN*                 & 81.52                           & 92.64 ($\uparrow$13.64\%)              & 85.59 ($\uparrow$4.99\%)               & 94.16 ($\uparrow$15.51\%)              & 81.22 ($\downarrow$0.37\%)               & 99.98 ($\uparrow$22.64\%)  & 99.71 ($\uparrow$22.31\%)   & 99.76 ($\uparrow$22.37\%) & 98.98 ($\uparrow$21.41\%)       & 99.95 ($\uparrow$22.61\%)       & 99.79 ($\uparrow$22.41\%)    \\
                              & LSUN                  & 77.22                           & 86.38 ($\uparrow$11.86\%)              & 76.58 ($\downarrow$0.83\%)             & 87.22 ($\uparrow$12.95\%)              & 77.54 ($\uparrow$0.41\%)                 & 63.67 ($\downarrow$17.54\%) & 62.31 ($\downarrow$19.30\%)  & 71.39 ($\downarrow$7.54\%) & 64.21 ($\downarrow$16.84\%)      & 61.47 ($\downarrow$20.39\%)      & 61.32 ($\downarrow$20.59)\%   \\ 
\hline
\multicolumn{2}{l|}{Avg \% improvement}  & \multicolumn{1}{l|}{}  & \multicolumn{1}{r}{($\uparrow$6.48\%)} & \multicolumn{1}{r}{($\uparrow$2.76\%)} & \multicolumn{1}{r}{($\uparrow$6.60\%)} & \multicolumn{1}{r}{($\downarrow$7.96\%)} &        \multicolumn{1}{r}{($\uparrow$5.15\%)}  &  \multicolumn{1}{c}{($\uparrow$4.98\%)}  &  \multicolumn{1}{r}{($\uparrow$6.26\%)}  &  \multicolumn{1}{c}{($\uparrow$4.82\%)}  &  \multicolumn{1}{c}{($\uparrow$4.80\%)}  &  \multicolumn{1}{l}{($\uparrow$4.90\%)}   \\
\hline
\end{tabular} %
}
\label{tab:real_data_aucroc}
\end{sidewaystable}

\clearpage
\begin{figure}[ht!]
    \centering
    \begin{subfigure}[t]{0.30\textwidth}
        \includegraphics[width=\textwidth]{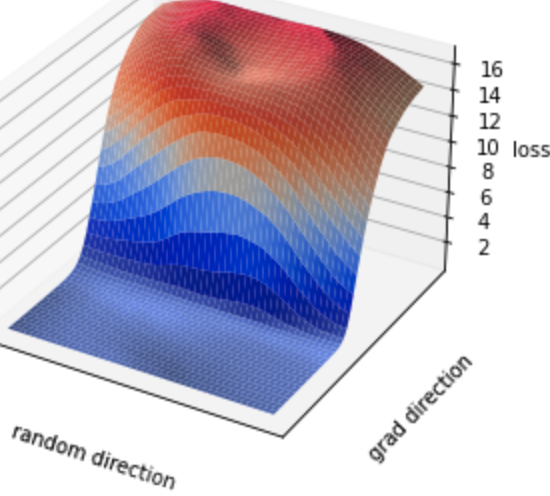}
        \caption{CrossEntropy}
    \end{subfigure}
    ~
    \begin{subfigure}[t]{0.30\textwidth}
        \includegraphics[width=\textwidth]{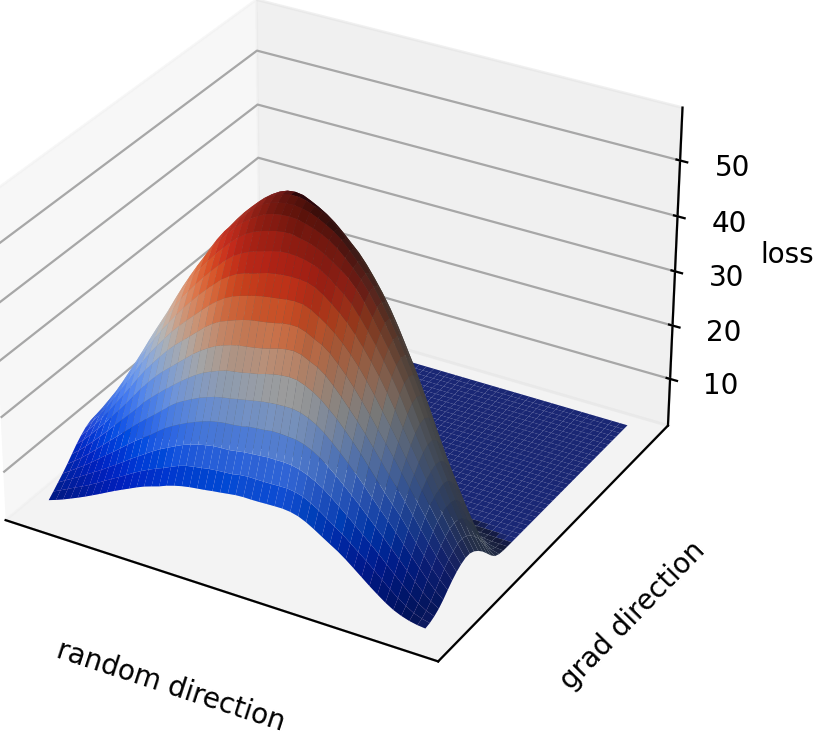}
        \caption{CrossEntropy+$\ell_{1}$}
    \end{subfigure}
     ~
    \begin{subfigure}[t]{0.30\textwidth}
        \includegraphics[width=\textwidth]{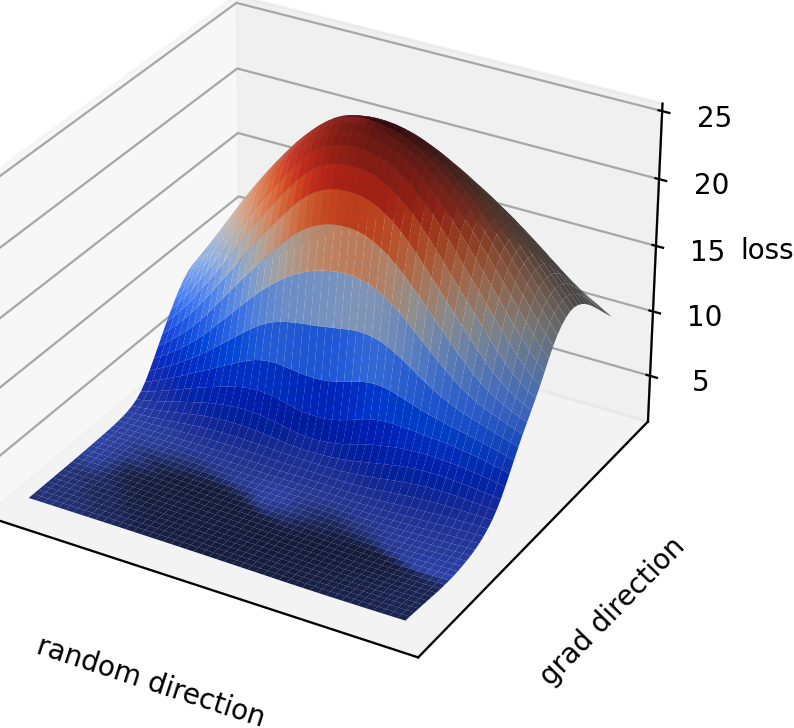}
        \caption{ContReg}
    \end{subfigure}
    
    \begin{subfigure}[t]{0.30\textwidth}
        \includegraphics[width=\textwidth]{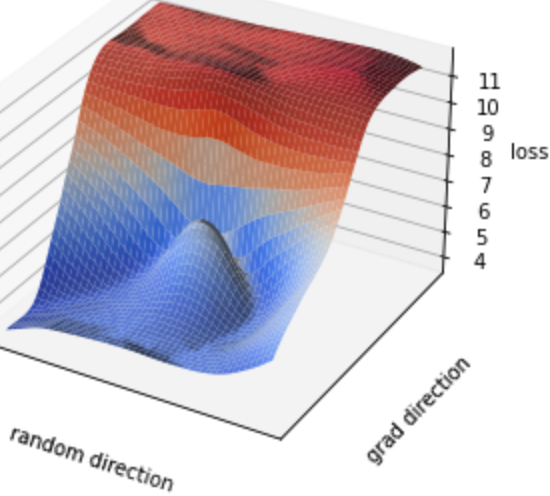}
        \caption{CrossEnt+MCD}
    \end{subfigure}
    ~
    \begin{subfigure}[t]{0.30\textwidth}
        \includegraphics[width=\textwidth]{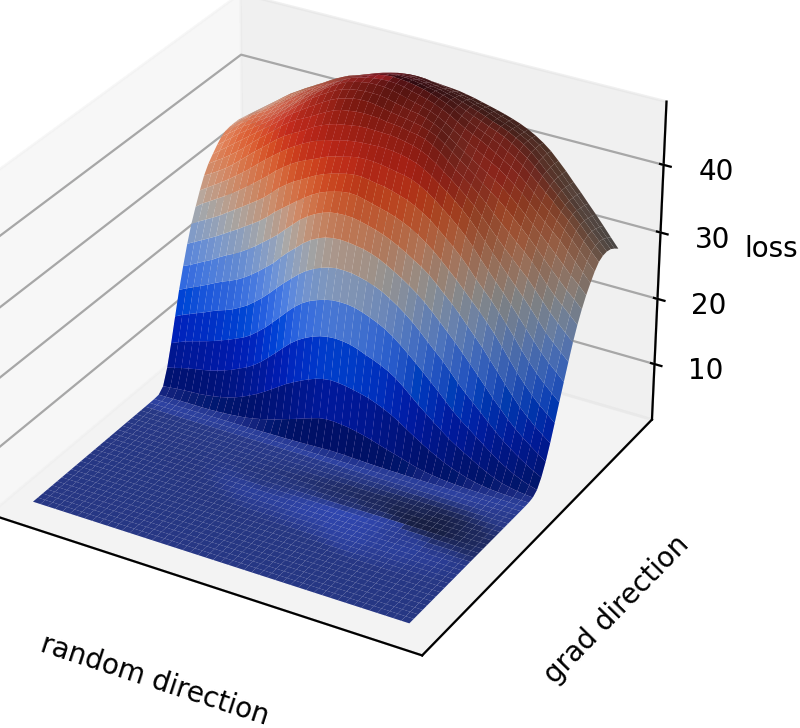}
        \caption{CrossEnt+$\ell_{1}$+MCD}
    \end{subfigure}
     ~
    \begin{subfigure}[t]{0.30\textwidth}
        \includegraphics[width=\textwidth]{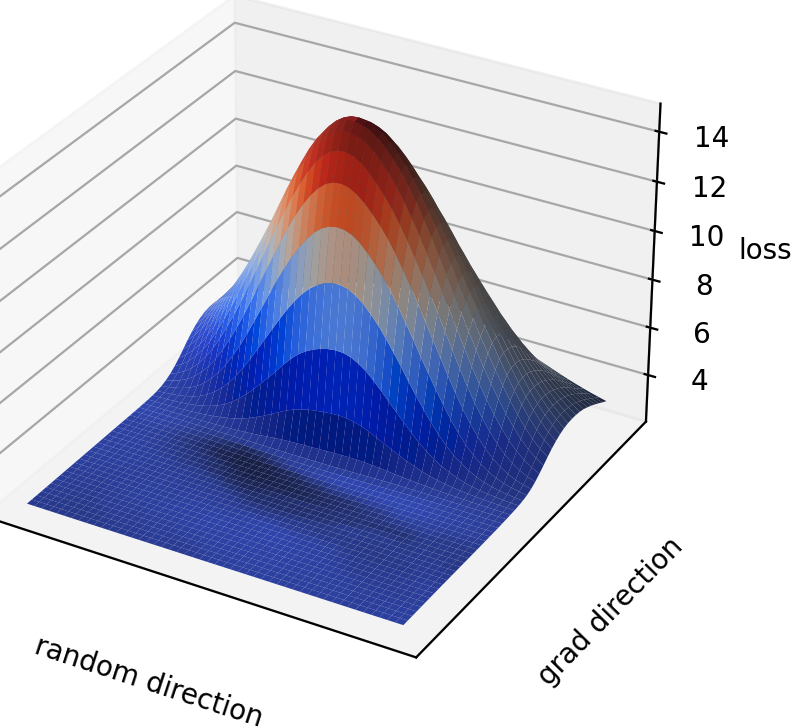}
        \caption{ContReg+MCD}
    \end{subfigure}
\caption{Comparison of different objectives trained on ID CIFAR-10 and tested on OOD CIFAR-100 with (1st row) and without (2nd row) explicit regularisation (MCD: Monte-Carlo Dropout).}
\label{fig:loss_projections}
\end{figure}
\section{Conclusion}
In this work we presented two novel objective functions with the goal of being utilised in a normal classification setting while at the same time exhibiting some robustness properties against common corruptions and ambiguous inputs when evaluated in OOD detection. We demonstrated that our approach outperforms half of the competitive methods and performs comparably to the remaining ones. Furthermore, we presented the efficacy of our method against common corruptions measured in mCE compared to competitive alternative methods. Finally, we identified the importance of auxiliary information as well as the role of regularisation in OOD detection, followed some important questions in identifying the role of bias in the choice of objective function, family class, and algorithm when considering open set classification problems.

\clearpage

\bibliographystyle{splncs04}
\bibliography{biblio}
\end{document}